\documentclass[11pt, a4paper]{article} 
\usepackage[margin=1.0in]{geometry}
\usepackage{graphicx}
\usepackage{url} 
\usepackage{algorithm}
\usepackage[noend]{algorithmic}
\usepackage{amsmath}
\usepackage{amssymb}

\newtheorem{theorem}{Theorem}
\newtheorem{proposition}{Proposition}[section]
\newcommand{\qed}{\hfill $\Box$}

\newenvironment{proofof}[1]{\par\noindent{\bf Proof of~\protect{#1}:}}{\qed \par\medskip}
\newcommand{\wh}{\widehat}
\newcommand{\spin}{\{-1,+1\}}

\newcommand{\norm}[1]{\left\|{#1}\right\|}

\newcommand{\bx}{\boldsymbol{x}}
\newcommand{\be}{\boldsymbol{e}}

\newcommand{\brf}{\bar{f}}
\newcommand{\brg}{\bar{g}}

\newcommand{\bd}{\boldsymbol{d}}

\newcommand{\loss}{\ell}

\newcommand{\yhat}{\wh{y}}

\newcommand{\bzero}{\boldsymbol{0}}
\newcommand{\bone}{\boldsymbol{1}}
\newcommand{\field}[1]{\mathbb{#1}}
\newcommand{\R}{\field{R}}

\newcommand{\scH}{\mathcal{H}}
\newcommand{\scA}{\mathcal{A}}

\newcommand{\scK}{\mathcal{K}}
\newcommand{\scS}{\mathcal{S}}

\newcommand{\tr}{\mbox{\sc trace}}
\newcommand{\rnd}{\mbox{\sc rnd}}

\newcommand{\argmin}{\operatornamewithlimits{argmin}}

\title{Memory Constraint Online Multitask Classification}

\author{
Giovanni Cavallanti \\
Department of Computer Science\\
Universit\`a degli Studi di Milano\\
Italy\\
\texttt{giovanni.cavallanti@unimi.it}\\
\\
Nicol\`o Cesa-Bianchi \\
Department of Computer Science\\
Universit\`a degli Studi di Milano\\
Italy\\
\texttt{nicolo.cesa-bianchi@unimi.it}
}

\begin{document} 

\maketitle

\begin{abstract} 
  We investigate online kernel algorithms which simultaneously process multiple
  classification tasks while a fixed constraint is imposed on the size of their
  active sets. We focus in particular on the design of algorithms that can
  efficiently deal with problems where the number of tasks is extremely high
  and the task data are large scale.  Two new projection-based algorithms are
  introduced to efficiently tackle those issues while presenting different
  trade offs on how the available memory is managed with respect to the prior
  information about the learning tasks.  Theoretically sound budget algorithms
  are devised by coupling the Randomized Budget Perceptron and the Forgetron
  algorithms with the multitask kernel.  We show how the two seemingly
  contrasting properties of learning from multiple tasks and keeping a constant
  memory footprint can be balanced, and how the sharing of the available space
  among different tasks is automatically taken care of. We propose and discuss
  new insights on the multitask kernel. Experiments show that online kernel
  multitask algorithms running on a budget can efficiently tackle real world
  learning problems involving multiple tasks.
\end{abstract} 

\section{Introduction}
In recent years there has been a growing interest in online learning
algorithms processing data from multiple and related sources. Many
interesting multitask problems involve large scale data sets or pose
memory and real-time restrictions. For example, massive personalized
spam detectors and ad serving systems need real-time, scalable, and
continuously adaptable learning methods.  Multi-sensors, memory
limited handheld devices deployed on the field are often required to
process and classify readings without relying on a centralized,
dedicated mainframe, and therefore face severe memory restrictions.
Online multitask algorithms are also a natural choice for a growing
number of applications that do not necessarily involve online data
processing, but where data sets are so large that the computationally
expensive batch algorithms can not be used (see, for example, \cite{BB08}).

In this paper we leverage on known results (\cite{CCG07},
\cite{DSS08}, \cite{OKC10}) to design online kernel algorithms that
effectively address these problems. First, we cast new light on how
the multitask kernel of~\cite{EMP05} acts as a proxy to wire prior
information about task relations into learning algorithms.  Second, we
build upon the Projectron algorithm~\cite{OKC10} to design two new and
highly efficient multitask budget online algorithms that
aggressively retain learned information by sharing the global budget across
different tasks. This is achieved using two multitask mediated projection
steps. We also show how existing budget online kernel
algorithms 
can be combined with the multitask paradigm in order to obtain
scalable and accurate solutions that retain strong theoretical bounds. In doing
so we show how the available space is automatically shared by and assigned to
different tasks.
Third, we provide an empirical evaluation of the proposed algorithms
in a variety of different experimental setups. We stress the fact that
our algorithms are particularly apt to be applied to problems where
both the data per task and the tasks themselves are great in
number. This goal is achieved by combining the mild dependence on the
number of tasks with the enforced budget size.

As of recently, several papers have considered large scale learning of
multitask problems. For example, \cite{WDLSA09} introduces a highly
scalable linear algorithm for spam classification based on hashing
techniques. Our work focuses on kernel algorithms, and therefore it
is not directly comparable to~\cite{WDLSA09}. Moreover, by relying on
the multitask kernel, the algorithms discussed here can be easily
applied to model situations where tasks exhibit a specific pattern of
relationships. A different approach is the one outlined
in~\cite{SRDV11}, where the focus is on learning and tracking
continuously changing relations among tasks. The nature of the problem
considered there makes their algorithms not suitable to large scale
applications, since the active set can easily grow unbounded and the
dependence on the number of tasks tends to be quadratic. In
Section~\ref{s:mtalgos} we show that under certain assumptions our
techniques can effectively deal with shifting tasks.

\section{Preliminaries and notation}
\label{s:premnot}
We consider the usual online classification protocol, where learning proceeds in
trials, with an additional complication due to the presence of multiple,
possibly related, tasks. At each time step $t$ an instance vector $\bx_t \in
\R^d$ for a given task $i_t \in \{1,\dots,k\}$, chosen among a fixed set of $k$
different tasks, is disclosed, and the algorithm outputs a corresponding binary
prediction $\yhat_t \in \spin$. The true label $y_t$ is then revealed and the
algorithm acts accordingly, choosing if and how to update its internal
state. We follow~\cite{CCG10} and define the $t$-th multitask instance as the
pair $[\bx_t, i_t]$. Similarly, the multitask example at time $t$ is defined to
be the pair comprised of the $t$-th multitask instance together with the label
$y_t$.  We do not assume any specific generative model for the sequence of the
multitask examples, i.e., no assumptions are made on the instance vectors
$\bx_1, \bx_2, \dots$, the task markers $i_1, i_2, \dots$ and the labels
$y_1, y_2, \dots$.  In this work our main interest focuses on classification
algorithms that are rotationally invariant and allow for the adoption of the
so-called kernel trick. Let $\scK: \left( \R^d \times \{1,\dots,K\} \right)
\times \left( \R^d \times \{1,\dots,K\} \right) \to \R$ be a symmetric, positive
semidefinite kernel operator between multitask instances and denote with
$\scH_{\scK}$ its associated Reproducing Kernel Hilbert Space (RKHS).
Following the online literature, theoretical results are stated in the form of
relative mistake bounds, where the number of mistakes made by the algorithm is
compared against a measure of performance obtained by the best reference
classifier in a given comparison class. We assume the standard hinge loss
function as measure of performance, which is defined, for any $\brg \in
\scH_{\scK}$, by
$
\loss_t(\brg) 
=
\max \left\{0, 1 - y_t \brg(\bx_t, i_t) \right\}
$.
Since theoretical results are given with respect to arbitrary sequences, we
also introduce the cumulative hinge loss $L(\brg)=\sum_t \loss_t(\brg)$.

With the intent of properly modeling a number of scenarios that are
frequently occurring in practice, we also consider the so-called
shifting model in which the reference classifier is allowed to change,
or shift, throughout the ongoing learning phase. Under this assumption
the single reference classifier $\brg$ is replaced by the sequence
$\brg_1, \brg_2, \dots$ and the cumulative hinge loss becomes
$L(\{\brg_t\}_{1,2,\dots})=\sum_t \loss_t(\brg_t)$. We omit the
argument of the cumulative loss whenever no ambiguity
arises. Unsurprisingly, the presence of shifting reference classifiers
will be reflected in theoretical bounds through a term that takes into
account the overall amount of shifting that the reference sequence
undergoes. Such term, known as the {\em total shift} and denoted with
$S_{\scA}(\{\brg_t\})$, or $S_{\scA}$ whenever
$\{\brg_t\}$ is understood from the context, is defined as
a sum of the distances computed with respect to a given positive
definite operator $\scA$ between consecutive classifiers in the
sequence, or, formally, as $S_{\scA}(\{\brg_t\})=\sum_t \norm{\scA(\brg_t - \brg_{t-1})}$.

As previously mentioned, this paper concentrates on kernel-based online linear
algorithms. The prediction function $\brf: \R^d \times \{1,\dots,K\} \to \R$ of
such algorithms can be encoded in the so-called dual form $\brf(\cdot)= \sum_j
\beta_j \scK \bigl([\bx_j, i_j], \cdot \bigr)$, that is, as a linear
combination of terms $\beta_j \scK\bigl([\bx_j, i_j], \cdot\bigr)$ where the
weights $\beta_j$'s are real coefficients.  We refer to the (multi)set of those
instances that appear in the linear combination for $\brf$ as to the {\em
  active set} $\scS$ and to the instances themselves as to active
instances. For convenience we also adopt the shorthand $J = \left \{j: [\bx_j,
  i_j] \in \scS \right\}$. A common drawback of the dual form expansion is the
tendency of the active set to grow unbounded. Online algorithms such that the
expanded or dual form is limited to $B$ terms are known as {\em budget}
algorithms. The budget requirement effectively imposes a memory constraint and
forces the online algorithm to throw away an active example whenever a new one
has to be added to the active set. In the rest of the paper we use the
expressions ``budget'' and ``active set'' interchangeably when we refer to
budget algorithms.

\section{From single to multiple tasks}
\label{s:sintomul}
We first provide a brief description of the multitask kernel as
introduced in~\cite{EMP05} and the rationale behind it.  Denote with
$\scK': \R^d \times \R^d \to \R$ the kernel operator between single
task instances and with $\scH_{\scK'}$ its associated RKHS.  In order
to streamline the presentation as much as possible we assume
$\norm{\bx_t}=\sqrt{\scK'(\bx_t, \bx_t)}=1$ for all $t$.

For our purposes the multitask kernel can be seen as a meta-kernel in that it
is responsible for {\em properly balancing} the impact that a given instance
$\bx_t$ has on the learning of task $i_t$ as well as of the possibly related,
remaining tasks $1,\dots,i_t - 1,i_t + 1,\dots,k$. Ideally, in order to meet
this goal the multitask kernel should be defined according to \textit{a priori}
information encoded in a graph that establishes mutual relations among
tasks. Let $G=(V_G,E_G)$ be such graph, with $V_G$ representing the tasks and
$E_G$ being their relations, and define the associated Laplacian matrix $L_G$
as
\[
(L_G)_{i,j} = \left\{
  \begin{array}{cl}
    d_i & \textrm{if $i=j$,}
    \\
    -1  & \textrm{if $(i,j) \in E_G$,}
    \\
    0   & \textrm{otherwise.}
  \end{array} 
\right.
\]
where $d_i$ denotes the number of tasks $i$ is related to.  Let $A_G=I+L_G$ be
the so-called {\em interaction matrix} for the graph $G$\footnote{The identity
  matrix ensures the positive definiteness of $A_G$ and allows for an easier
  treatment. It is however possible to define $A_G=L_G$ and then use the
  pseudoinverse $A_G^{+}$ in place of the inverse $A_G^{-1}$.}.  The graph
induced kernel product is defined by $ \scK
\bigl([\bx_s,i_s],[\bx_t,i_t]\bigr)=(A_G^{-1})_{i_s, i_t} \scK'(\bx_s,\bx_t) $.
Given a sequence of multitask examples, it easily follows that $\sqrt{\scK
  \bigl([\bx_s,i_s],[\bx_t,i_t]\bigr)} \le \max_{i}
(I+L_G)^{-1/2}_{i,i} = c_G$ for all time steps $s$ and $t$ in the sequence.
The magnitude of $c_G$ scales according to the connectivity of the multitask
kernel inducing graph $G$. In particular, $c_G$ ranges from $\sqrt{2/(k+1)}$ if
$G$ is complete to up to $1$ when $G$ has at least one isolated task.
Note that, since the model considered here involves multiple tasks, it
is often found to be more natural to actually describe any reference
classifier as set of classifiers, each one representing a classifier
for a subsequence $\bx_{t_1}, \bx_{t_2}, \dots$ where $t_1,t_2,\dots$
are such that $i_{t_1}=i_{t_2}=\dots$. For the sake of convenience we
adopt an extended, vector-based notation and denote with $\brg =
\bigl[g_1,\dots,g_k \bigr]$ the multitask reference classifier made by
the single task classifiers $g_1 \in \scH_{\scK'}, \dots, g_k \in
\scH_{\scK'}$.  Moreover, by denoting with $\bzero(\cdot)$ the null
operator in $\scH_{\scK'}$ and keeping in with the above notation, it
turns out that the kernel inducing feature map $\psi_{\scK}$ is such
that $\psi_{\scK}\bigl([\bx_t,i_t]\bigr) =
\scK\bigl([\bx_t,i_t],\cdot\bigr) =
\bigl[\bzero(\cdot),\dots,\bzero(\cdot),
\scK'(\bx_t,\cdot),\bzero(\cdot),\dots,\bzero(\cdot) \bigr] A_G^{-1/2}
$, therefore mapping multitask instances to the space of vector valued
functions endowed with the inner product $\langle \brf, \brg
\rangle=\tr(K_{\brf, \brg})$, where $(K_{\brf, \brg})_{i,j}=\langle
f_i, g_j \rangle$ for any $\brf, \brg \in \scH_{\scK}$ is the standard
inner product for the space $\scH_{\scK'}$.  It is also worthwhile to
observe that, when the multitask kernel is employed, and the foregoing
interpretation on the structure of $\brg \in \scH_{\scK}$ holds, the
hinge loss incurred at time $t$ by $\brg$ reduces to the hinge loss
incurred on the example $(\bx_t, y_t)$ by the $i_t$-th single task
reference classifier, i.e, $\loss_t(\brg)=\max \left\{0, 1 - y_t
  g_{i_t}(\bx_t) \right\}$. %
The multitask kernel is therefore a meta-kernel since it is a scalar multiple
of the underlying kernel computed on the single task instances,
disregarding task relations. This latter information, indeed a pair of
coordinates pointing to an entry in the inverse of the interaction
matrix $A_G$, is instead taken into account to determine the actual
value of the scaling factor.
In particular, the following statement holds (proof deferred to the
Appendix).
\begin{proposition}
  \label{pr:rescoeff}
  Let $G$ be a graph of $k$ nodes and $G'$ be the augmented graph
  obtained from $G$ by adding a dummy node that is connected to every
  nodes in $G$. Then $\bigl(A_G^{-1}\bigr)_{i,j}$ is equal to
\[
  -\frac{1}{2}(R_{G'})_{i,j}+\frac{1}{2(k+1)}\sum_{l=1}^{k+1}(R_{G'})_{i,l}
+\frac{1}{2(k+1)}\sum_{l=1}^{k+1}(R_{G'})_{j,l} - \frac{1}{(k+1)^2}\norm{R_{G'}}_1 +\frac{k+2}{(k+1)^2}
\]
where $\norm{\cdot}_1$ denotes the entrywise $1$-norm and $R_{G'}$ is
the resistance matrix whose entries $(R_{G'})_{i,j}$ are the
resistance distances between tasks $i$ and $j$ in the augmented graph
$G'$.
\end{proposition}
The augmented graph is needed due to the fact that $A_G = I + L_G$ and
therefore it is not a Laplacian matrix.  Note %
that $(A_G^{-1})_{i,i}$ is $1$ if task $i$ is isolated\footnote{It is enough to
  observe that the entries on $i$-th row and the $i$-th column are zeros.} and
gets smaller and smaller as the number of its related tasks grows.  On the
other hand, $(A_G^{-1})_{i,j}$, with $i \neq j$, decreases as the resistance
distance between tasks $i$ and $j$ grows, and as the average resistance
distances between each of the two tasks $i$ and $j$ and the rest of the tasks
in the graph increase.
This amounts to say that two identical single task instances from
loosely connected tasks may be considered more ``different'' than
slightly different single task instances from tightly connected
tasks. The magnitude of $(A_G^{-1})_{i,i}$ taken in isolation has more
a balancing role than anything else.
In particular, note that $(A_G^{-1})_{i,j}$ for $i=j$ is always greater than
$i\neq j$ for all tasks $i$,$j$ belonging to the same connected component. Indeed,
as we expect, $\scK \bigl([\bx_t,i],[\bx_t,i]\bigr) \ge \scK \bigl([\bx_t,i],[\bx_t,j]\bigr)$ for any $j \neq i$. 
Finally, it is worth mentioning that, while some of the more compelling
properties of the multitask kernel arise from its interpretation in terms of
the graph $G$, nothing prevents one to replace the Laplacian $L_G$ with an
arbitrary positive semidefinite matrix.

\section{Dealing with multiple tasks with a memory budget constraint}
\label{s:mtalgos}
Motivated by the elegant and general theoretical guarantees and by the easiness
of their implementation, we now introduce several budget algorithms for the
multitask setup.
\subsection{The Multitask Budget Projectron}
The first multitask algorithm (Algorithm~\ref{a:mtBudgetProjectron}, {\sc
  mtbprj}) considered here is a modification of the Projectron
algorithm~\cite{OKC10} where a hard constraint is imposed on the size of the
active set.  Similarly to~\cite{WCV10}, the general idea of the algorithm aims
at optimizing the allotted space by capitalizing on the examples already stored
in the active set.
\begin{algorithm}[H]
  \caption{{\sc mtbprj}} 
  \label{a:mtBudgetProjectron} 
  \begin{algorithmic}[1]
    \REQUIRE Graph $G$, Budget size $B > 0$, Projection threshold $\eta>0$
    \STATE $\scS \leftarrow \emptyset$ \FORALL{$t=1,2,\dots$}

    \STATE Get $\bigl([\bx_t, i_t], y_t \bigr)$ and let $f_{i_t}(\bx_t)$ be $\sum_{j \in J} \beta_j (A_G)^{-1}_{i_j, i_t}
    \scK'(\bx_j, \bx_t)$

    \IF{$ y_t f_{i_t}(\bx_t) \le 0$}

    \IF{$\norm{P_J^{\perp} \scK\bigl([\bx_t, i_t], \cdot\bigr)} \le \eta$}
    
    \STATE $\beta_j \gets \beta_j + y_t \alpha_j$, $\forall j \in J$
    \{see text\}%

    \ELSE

    \STATE $\beta_t \gets y_t$

    \IF{$|\scS| \le B$}

    \STATE  $\scS \gets \scS \cup [\bx_t, i_t]$ 

    \ELSE 

    \STATE $r \gets \argmin_{j \in J} \beta_j \norm{P_{J \setminus \{j\} \cup
        \{t\}}^{\perp} \scK\bigl([\bx_j,i_j], \cdot\bigr)}$

    \STATE $\scS \gets \scS \cup [\bx_t, i_t] \setminus [\bx_r, i_r]$

    \STATE $\beta_j \gets \beta_j + \beta_r \gamma_j$, $\forall j \in J$ \{see text\}

    \ENDIF

    \ENDIF 

    \ENDIF

    \ENDFOR
  \end{algorithmic} 
\end{algorithm}
Denote with $P_J(\cdot)$ the projection operator on the space spanned by
$\{\scK\bigl([\bx_j,i_j],\cdot \bigr)\}_{j \in J}$ and with
$P_J^{\perp}(\cdot)$ the corresponding orthogonal projection operator. When a
mistake occurs at time step $t$, the expression $ \norm{P_J^{\perp} \bigl(
  \scK\bigr([\bx_t, i_t], \cdot\bigl) \bigr)} =
\norm{P_J\left(\scK\bigl([\bx_t,i_t],\cdot\bigr) \right) - \scK\bigl([\bx_t,
  i_t],\cdot\bigr)} $ is evaluated as way to assess how much of
$\scK\bigl([\bx_t, i_t], \cdot\bigr)$ can not be written as a linear
combination of the current active multitask instances.
In particular, if $\norm{P_J^{\perp} \bigl( \scK\bigl([\bx_t, i_t], \cdot
  \bigr) \bigr)}$ is smaller than the user supplied threshold $\eta$ (i.e., a
portion of size no bigger than $\eta$ will not be preserved after projection),
then the budget is left untouched and the weights $\beta_j$'s are updated to
reflect that we are actually storing
$P_J\bigl(\scK\bigl([\bx_t,i_t],\cdot\bigr)\bigr)$ in place of
$\scK\bigl([\bx_t,i_t],\cdot \bigr)$. In fact,
$P_J\bigl(\scK\bigl([\bx_t,i_t],\cdot\bigr)\bigr) = \sum_{j \in J} \alpha_j
\scK\bigl([\bx_j,i_j],\cdot\bigr)$ where the $\alpha_j$'s are the entries of
the vector $H^{-1} \bigl[\cdots \scK\bigl([\bx_l,i_l],[\bx_t, i_t] \bigr)
\cdots\bigr]^{\top}$ , $\forall l \in J$, and $H$ denotes the Gram matrix of
the current active multitask instances.  The projection step depends on task
relations in such a way that the condition on line 6 is unlikely to be true if
$i_t$ is loosely connected to tasks $i_l$, even if $\scK'(\bx_t, \cdot)$ is in
the space spanned by $\{\scK'(\bx_j,\cdot)\}_{j \in J}$ . Otherwise, either
$|\scS| \le B$ and $\scK\bigl([\bx_t,i_t],\cdot \bigr)$ is simply loaded into
the budget or $|\scS| = B$ in which case a projection-based budget maintenance
policy is triggered. As result, {\sc mtbprj} singles out for eviction the
multitask instance $[\bx_r, i_r]$ that can be removed and projected back onto
the remaining active instances with little overall damage as measured by
$\beta_r \norm{P_{J \setminus \{r\} \cup \{t\}}^{\perp} \scK\bigl([\bx_r,i_r],
  \cdot\bigr)}$. Here $\norm{P_{J \setminus \{r\} \cup \{t\}}^{\perp}
  \scK\bigl([\bx_r,i_r], \cdot\bigr)}$ is the amount of $\scK\bigl([\bx_r,i_r],
\cdot\bigr)$ that is lost after projecting it back. The weights $\beta_j$'s are
then updated pretty much in the same way as on line $7$ with $\gamma_j$'s %
being the coefficients of the expansion of $\scK\bigl([\bx_r,i_r], \cdot\bigr)$
as a linear combination of $\bigl\{\scK\bigl([\bx_j,i_j],\cdot\bigr)\bigr\}_{j
  \in J \setminus \{r\} \cup \{t\}}$.

Unfortunately, {\sc mtbprj} suffers a major drawback in that instances from an
isolated task tend to undermine the projection mechanism. In fact, if tasks
$i_t$ and $i_j$'s are unrelated, $\scK([\bx_t,i_t],[\bx_j,i_j])=0$ which
implies $\norm{P_J^{\perp} \scK([\bx_t, i_t], \cdot) } > \eta$ almost
surely. This in turn prevents the efficient storage of $\scK\bigl([\bx_t, i_t],
\cdot \bigr)$ in terms of its projection.  For the same reason active multitask
instances from isolated tasks are seldom chosen for eviction since no part of
them can be retained as a linear combination of the remaining instances.
\begin{algorithm}[t]
  \caption{{\sc mtbprj-2}}
  \label{a:mtBudgetProjectron2} 
  \begin{algorithmic}[1]
    \REQUIRE Graph $G$, Budget size $B > 0$, Projection threshold $\eta > 0$
    \STATE $\scS \leftarrow \emptyset$ \FORALL{$t=1,2,\dots$}

    \STATE Get $\bigl([\bx_t, i_t], y_t \bigr)$ and let $f_{i_t}(\bx_t)$ be $\sum_{j \in J} (\beta_{i_t})_j
    \scK'(\bx_j, \bx_t)$

    \IF{$ y_t f_{i_t}(\bx_t) \le 0$}

    \IF{$\norm{(P')_J^{\perp} (\scK'(\bx_t,\cdot)) } \le \eta$}

    \STATE $(\beta_{l})_j \gets (\beta_l)_j + \alpha_j y_t (A_G^{-1})_{i_l,
      i_t}$, $\forall j \in J, l \in \{1,\dots,k\}$

    \ELSE

    \IF{$|\scS| \le B$}

    \STATE  $\scS \gets \scS \cup [\bx_t, i_t]$ 

    \STATE $(\beta_{l})_t \gets y_t (A_G^{-1})_{i_l,i_t}$, $\forall l
    \in \{1,\dots,k\}$

    \ELSE

   \STATE $r \gets \argmin_{j \in J} \norm{\bd_j}$ \{see text\}

    \STATE $\scS \gets \scS \cup [\bx_t, i_t] \setminus [\bx_r, i_r]$

    \STATE $(\beta_{l})_j \gets (\beta_l)_j + \gamma_j (\beta_l)_r
    (A_G^{-1})_{i_l, i_j}$, $\forall j \in J, l \in \{1,\dots,k\}$

    \ENDIF 

    \ENDIF

    \ENDIF

    \ENDFOR
  \end{algorithmic} 
\end{algorithm}
To overcome these limits we designed a new projection-based budget
multitask algorithm (Algorithm~\ref{a:mtBudgetProjectron2}, {\sc
  mtbprj-2}).  We denote with $P'_J(\cdot)$ the projection operator on
the space spanned by $\{\scK'(\bx_j,\cdot)\}_{j \in J}$ and with
$(P')^{\perp}_J(\cdot)$ the corresponding orthogonal projection
operator. 
The algorithm maintains $k$ sets of weights $(\beta_i)_j$'s for $i=1,\dots,k$
and in doing so it trades off space (used to store the weights $(\beta_i)_j$)
to increase retention and ultimately accuracy. %
More specifically, the projection applied to the instance observed at time $t$
is equivalent to the one employed by {\sc mtbprj} except that the task markers
are now ignored, effectively increasing the chance of the instance to be
efficiently stored through its projection. The projection based maintenance
policy now prescribes to remove an active instance $\scK'(\bx_j,\cdot)$ so that
the $l_2$-norm of the vector $ \bd_j= \left[ (\beta_1)_j \norm{(P')^{\perp}_{J
      \setminus \{j\} \cup \{t\}} \scK'(\bx_j,\cdot)}\ \cdots \ (\beta_k)_j
  \norm{(P')^{\perp}_{J\setminus \{j\} \cup \{t\}} \scK'(\bx_j,\cdot)} \right]
$
is as small as possible.  As one can easily observe, each of the $k$ entries of
this vector gauges to what extent the removal of $\scK'(\bx_j,\cdot)$ affects
one of the different $k$ prediction functions that the algorithms maintains.
\subsection{The Multitask Randomized Budget Perceptron}
\label{ss:mtRBP}
The next algorithm we consider (Algorithm~\ref{a:mtRBP}, {\sc mtrbp}) is the
multitask version of the Randomized Budget Perceptron~\cite{CCG07} algorithm.
\begin{algorithm}[H]
  \caption{{\sc mtrbp}}%
  \label{a:mtRBP}
  \begin{algorithmic}[1]
    \REQUIRE Graph $G$, Budget size $B > 0$
    \STATE $\scS \leftarrow \emptyset$ 
    \FORALL{$t=1,2,\dots$}

    \STATE Get $\bigl([\bx_t, i_t], y_t \bigr)$ and let $f_{i_t}(\bx_t)$ be $\sum_{j \in J} \beta_j (A_G)^{-1}_{i_j, i_t}
    \scK'(\bx_j, \bx_t)$

    \IF{$ y_t f_{i_t}(\bx_t) \le 0$}
    
    \STATE $\beta_t \gets y_t$
    
    \IF{$|\scS| < B$}

    \STATE $\scS \gets \scS \cup [\bx_t, i_t]$

    \ELSE 

    \STATE $\scS \gets \scS \cup [\bx_t, i_t] \setminus \rnd(\scS)$

    \ENDIF 

    \ENDIF

    \ENDFOR
  \end{algorithmic} 
\end{algorithm}
{\sc mtrbp} resembles the Perceptron %
algorithm and relies on a very simple scheme to deal with the memory budget
constraint. If a mistake occurs on time step $t$, the instance $[\bx_t, i_t]$
is added to the budget with the weight set to $y_t$ and a random-based space
management policy is triggered whenever the budget size grows beyond $B$.  In
the pseudocode of Algorithm~\ref{a:mtRBP} the primitive $\rnd(\cdot)$ samples a
random element from its set argument.
The following theorem is a fairly straightforward multitask version
of~\cite[Theorem 5]{CCG07}. Since the algorithm is randomized, the theoretical
guarantee provided in Theorem~\ref{t:mtRBP} bounds the expected value of the
(random) number of mistakes, rather than the number of mistakes itself.
\begin{theorem}
  \label{t:mtRBP}
  The expected value of the number of mistakes $M$ made by the {\sc
    mtrbp} algorithm, run with a graph $G$ and with a budget size
  $B>0$, on any finite sequence of $n$ multitask examples $\bigl(
  [\bx_1, i_1], y_1 \bigr), \dots , \bigl( [\bx_n, i_n], y_n \bigr)$,
  is bounded above by
  \begin{equation}
    \frac{1}{1-\epsilon} \left( \vphantom{\frac{\epsilon B^{3/2}}{\sqrt{2}}}
      L
      + c_G
      S_{A_G^{1/2}}\sqrt{B} %
      +
      \frac{\epsilon B^{3/2}}{2}
      + 
      \frac{\epsilon B}{4} \ln \frac{B}{3}
    \right)
    \label{e:mtRBPbnd}
  \end{equation}
  for any $\epsilon \in (0,1)$ and any sequence of multitask reference
  classifiers $\brg_0,\dots,\brg_{n-1} \in \scH_{\scK}$ such that $\max_t
  \sqrt{ \tr (K_{\brg_t, \brg_t} A_G) } \le \frac{1}{2 c_G} \epsilon \sqrt{B}$
  where $K_{\brg_t, \brg_t}$ is the Gram matrix computed among the single task
  classifiers $(g_t)_1, \dots, (g_t)_k$.
\end{theorem}
The variable $\epsilon$ in Theorem~\ref{t:mtRBP} trades off the size of the
comparison class for the tightness of the bound, the larger the former (which
is the case when $\epsilon$ leans towards $1$), the looser the latter. Since
Theorem~\ref{t:mtRBP} holds for arbitrary choices of $\epsilon$, {\sc mtrbp}
effectively competes against the best multitask classifier in the best traded
off comparison class.  Let us observe how the multitask kernel affects
bound~(\ref{e:mtRBPbnd}).  The role of the kernel appears evident in the
shifting term, both explicitly through the presence of the interaction matrix
$A_G$ and implicitly through the weighting constant factor $c_G$. Consider the
expansion of $S_{A_G^{1/2}}$ as $\sum_{t=1}^{n-1} \sqrt{\bigl \langle
  (\brg_t-\brg_{t-1})A_G, (\brg_t-\brg_{t-1}) \bigr\rangle }$. Each term under
the square root has the form
\begin{equation}
\label{e:mtRBPshift}
\sum_{i=1}^k \norm{g_{t,i} - g_{t-1,i}}^2 + \sum_{(i,j) \in E}
\norm{(g_{t,i} -g_{t,j}) - (g_{t-1,i}-g_{t-1,j})}^2
\end{equation}
where the first summand summarizes how much each of the $k$ reference
classifiers shifts from time step $t-1$ to time step $t$ and therefore does not
take into account the relations among tasks. The second summand
of~(\ref{e:mtRBPshift}) is where those relations come into play. Specifically,
for each pair of related tasks the difference of their relative positions after
consecutive time steps is evaluated. This expression is clearly small when the
reference classifiers of related tasks shift in similar ways\footnote{Think of
  a multi language spam classifier. Different trends may arise over time but
  the language(task) relations stay the same.}. In particular, when this
shifting pattern holds and $G$ is a complete graph the shifting term $ c_G
S_{A_G^{1/2}}$ becomes, excluding constant factors, $ \sum_{t=1}^{n-1}
\sqrt{\frac{\sum_{i=1}^k \norm{g_{t,i} - g_{t-1,i}}^2}{k+1}} $ and it is
therefore similar to the one we would get if there were only one task.  Aside
from the shifting term, the impact of the multitask kernel is then largely
confined in the comparison class inequality that binds $B$ to $\tr
(K_{\brg_t,\brg_t} A_G)$. In fact, setting a given value of $B$ amounts to
define the shape of the class of multitask reference classifiers the algorithm
competes against. After rewriting $\tr (K_{\brg_t, \brg_t} A_G)$ as $
\sum_{i=1}^K \norm{g_{t,i}}^2 + \sum_{(i,j) \in E} \norm{g_{t,i} - g_{t,j}}^2$
it is easy to see that a given choice of $B$ imposes a constraint on the norms
of the task-specific reference classifiers {\em and} on the spatial relations
they entertain with each other. To better illustrate how the memory constraint
works in this respect, consider the following two opposite situations where we
again set $G$ to be the complete graph. First assume that the worst-case
multitask reference classifiers are stationary, i.e., $\brg_t = \brg$ for all
$t$, and their single task classifiers are overlapped, i.e.,
$g_1=g_2=\dots=g_k$. In this case $\norm{g_i}^2 \le \frac{k+1}{k}B$ for all
$i=1,\dots,k$, excluding constant factors. In other words, the algorithm can
compete against longer reference classifiers whose norm is nearly $B$ instead
of $B/k$ as a naive, non multitask approach would imply.  Implementation-wise,
this means that the whole allotted space can be devoted to learn a single, more
complex unique task rather than inefficiently fragmented to track $k$ equal
reference classifiers. On the other hand, if the $k$ single task reference
classifiers are distant from each other\footnote{When the reference classifiers
  have the same norm and $G$ is a complete graph this amounts to say that the
  $(g_t)_i$'s are the vertices of a $k$-simplex centered at the origin.}, as
measured by the metrics defined by $A_G^{1/2}$, then the average norm of the
single task reference vectors the algorithm can compete against is reduced by
an amount proportional to how much they are spread apart. In this case, since
the tasks we are learning are different from each other, {\sc mtrbp} is forced
to reserve a portion of the available space to each task.
\subsection{The Multitask Self-Tuned Forgetron}
Adapting the Forgetron algorithm of \cite{DSS08} to run within the multitask
protocol is relatively straightforward. Before investigating the details of the
algorithm ({\sc mtforg}) recall that its theoretical behavior is sub-optimal,
since the size of the comparison class is of the order $O(\sqrt{B/\log(B)})$, a
factor $\sqrt{\log(B)}$ worse than the optimal $O(\sqrt{B})$ achieved by {\sc
  mtrbp}\footnote{For details on why $O(\sqrt{B})$ is the largest norm an
  algorithm with a budget size $B$ can compete against
  see~\cite{DSS08}.}. Nonetheless, the algorithm may prove to be more effective
in a number of real world scenarios where the sequence of examples is not %
adversarial %
and the policy of always dropping the oldest instance may turn up to be
beneficial.

We consider the self-tuned version of the algorithm which performs
Perceptron-like updates whenever there is still room in the active set and
operates as follows otherwise. %
If a mistake occurs at time step $t$ and the current budget size $|\scS|$ is
already $B$, then the oldest active multitask instance $[\bx_r, i_r]$ with
$r=\min(J)$ is singled out for eviction and the incoming instance $[\bx_t,
i_t]$ is loaded into the budget with the weight $\beta_t$ set to $y_t$ (removal
step). As a way to control the detrimental effect of the removal of $[\bx_r,
i_r]$ from the budget, {\sc mtforg} also reduces the weights $\beta_j$'s by an
adaptive factor $\phi$ (shrinking step) so that older instances have smaller
weights. Because it is likely that both steps negatively affect the overall
performance of the algorithm, the rescaling factor $\phi$ is adaptively set to
moderate the impact of the shrinking step. We take advantage of the fact that
the norm of multitask instances is bounded by $c_G$, which may be much less
then $1$ for several non trivial graphs, and slightly fine tune the algorithm
presented in~\cite{DSS08} by setting
\begin{equation}
\label{e:phidef}
\phi = \max_{\chi \in (0,1]}\biggl(\Psi_G(\beta_r y_r \chi, \beta_r \chi
f_{i_r}(\bx_r)) + Q \le \frac{15 c_G^2 M}{32}\biggr)
\end{equation}
where $ \Psi_G(\lambda, \mu)=c_G^2 \lambda^2 + 2 c_G \lambda - 2
\lambda \mu $.  The resulting algorithm is similar to
Algorithm~\ref{a:mtRBP} where line~9 is replaced by
\[
\begin{array}{rl}
  9: & r \gets \min(J) \\
  10: & \scS \gets \scS \cup [\bx_t, i_t] \setminus [\bx_r, i_r] \\
  11: & \beta_j \to \phi \beta_j, \forall j \in J \{\phi \text{ computed as in (\ref{e:phidef})}\} \\
  12: &  Q \gets Q + \Psi_G(\beta_r y_r \phi, \beta_r \phi  f_{i_r}(\bx_r))
\end{array}
\]
The following theorem, which is an easy consequence of~\cite[Theorem 3]{DSS08}, provides a deterministic worst-case upper bound on the
number of mistakes made by {\sc mtforg}.
\begin{theorem}
  The number of mistakes $m$ made by the {\sc mtforg} algorithm, run
  with a graph $G$ and with a budget size $B>83$, on any finite
  sequence of $n$ multitask examples $\bigl([\bx_1, i_1], y_1\bigr),
  \dots, \bigl([\bx_n, i_n], y_n\bigr)$ satisfies
  \[
  m \le 4 L + \frac{B+1}{2 \log(B+1)}
  \]
  for any multitask reference classifier $\brg$ such that $\sqrt{ \tr
    (K_{\brg,\brg} A_G) } \le \frac{1}{4c_G}\sqrt{\frac{B+1}{\log(B+1)}}$ and
  $K_{\brg, \brg}$ is the Gram matrix computed among the single task
  classifiers $g_1, \dots, g_k$.
\end{theorem}
First, observe that the above bound only applies to a stationary comparison
multitask classifier. As far as we know no shifting analysis is known for the
Self-Tuned version of the Forgetron algorithm. Nonetheless, a shifting bound
can be obtained for an experimentally less appealing variant by following the
argument and the analysis given in~\cite{S09}. Second, as for {\sc mtrbp}, the
role of the multitask kernel is mainly reflected in the bound through the
comparison class inequality $\sqrt{ \tr (K_{\brg,\brg} A_G) } \le
\frac{1}{4c_G}\sqrt{\frac{B+1}{\log(B+1)}}$. In this respect, note that it is
the presence of $c_G$ in~(\ref{e:phidef}) that allows the size of the
comparison class to scale as a function of the tasks through the constant
factor $1/4c_G$ (see subsection~\ref{ss:mtRBP} for a detailed discussion on the
comparison class inequality within the multitask framework). Third, observe
that when a mistake occurs and $|\scS|$ is $B$, then the removal step only
affects those tasks that are related to $i_r$, since
$\scK([\bx_r,i_r],[\bx_j,i_j])=0$ if tasks $i_r$ and $i_j$'s are unrelated. As
a result, tasks belonging to connected components that seldom need processing
can be quickly forgotten altogether. Combining the Forgetron budget maintenance
policy with the multitask kernel has thus the important effect that the
allocation of the available space to the most frequent tasks is automatically
taken care of.
\subsection{Implementation details}
While the implementation of the budget algorithms discussed in this paper is
straightforward, it is still useful to point out a few remarks. Except for {\sc
  mtprj-2}, all the algorithms discussed in this paper only maintain $B$ real
weights and $B$ multitask instance vectors {\em regardless} of the number $k$
of the tasks at hand. Of course the matrix $A_G$ employed by the multitask
kernel may still require $O(k^2)$ space. Note, however, that if $G$ is not
overly complex and exhibits a certain regularity, the required space may be
much smaller, and of course it may be more efficient to opt for a programmatic
implementation of $A_G$ over the naive table-based approach. Both {\sc mtrbp}
and {\sc mtforg} require $O(1)$ operations to update their internal state on
mistaken rounds, whereas it is not hard to show that {\sc mtprj} take $O(B^2)$
operations. As for {\sc mtprj-2} we should note that the algorithm only needs
to store $B$ real weights for each connected component of $G$ which results in
a much milder dependence on $G$ than a naive implementation would imply.
\section{Experiments}
\label{s:exps}
\begin{table*}[t!]
  \caption{Online training F-measures achieved by the 
    multitask budget algorithms run with G 
    set to C (complete graph) or D (totally disconnected graph) and B set 
    to $25\%$, $10\%$ or $5\%$ of the size of the active set obtained by
    a battery of $k$ Perceptrons on the 
    PKDD 2006 spam task A, School binary, and Sentiment data sets. As a reference, the
    F-measures achieved by the $k$ independent Perceptrons
    are shown below the name of each data set.}
  \label{t:exps}
\vskip 0.15in
\begin{center}
\begin{small}
\begin{sc}
\begin{tabular}{l|l|cc|cc|cc}
  \hline
  & algorithm & C($25\%$) & D($25\%$) & C($10\%$) & D($10\%$) & C($5\%$) & D($5\%$) \\
  \hline
  {\bf Pkdd06 Spam A} & mtrbp & 84.8\% & 81.4\%  & 78.0\%  & 73.1\%  & 72.8\%  & 66.7\%  \\
  (92.0\%) &mtforg & 85.1\% & 81.7\%  & 77.9\%  & 73.0\%  & 72.3\%  & 66.7\% \\
  &mtbprj & 91.6\% & 89.4\%  & 87.8\% & 84.0\%  & 83.4\%  & 76.0\% \\
  &mtbprj-2 & 91.6\%  & 90.3\%  & 89.3\%  & 87.3\%  & 85.4\%  & 82.3\%  \\
  \hline
  {\bf School binary} &mtrbp & 40.4\% & 35.0\% & 38.6\% & 30.9\% & 37.3\% & 26.0\% \\
  (39.1\%)&mtforg & 39.7\% & 35.1\% & 38.0\% & 31.5\% & 36.9\% & 25.9\% \\
  &mtbprj & 40.6\% & 37.4\% & 40.2\% & 32.6\% & 39.4\% & 23.8\% \\
  &mtbprj-2 & 41.2\% & 39.1\% & 40.9\% & 39.0\% & 39.6\% & 37.9\% \\
  \hline
  {\bf Sentiment} &mtrbp & 66.8\% & 63.3\% & 62.0\% & 58.3\% & 59.4\% & 56.1\% \\
  (71.5) &mtforg & 66.7\% & 63.4\% & 62.3\% & 58.8\% & 58.8\% & 56.5\% \\
  &mtbprj & 71.4\% & 67.7\% & 66.6\% & 63.6\% & 64.0\% & 60.6\% \\
  &mtbprj-2 & 71.7\% & 68.5\% & 67.9\% & 64.9\% & 64.5\% & 61.7\% \\
  \hline
\end{tabular}
\end{sc}
\end{small}
\end{center}
\end{table*}
The experimental performance of budget multitask algorithms on non-synthetic
data sets is of key importance because the constraint on the size of the budget
and the favorable dependence on the number of tasks make them %
suited even for large scale applications. In this section we evaluate the
multitask budget algorithms discussed in Section~\ref{s:mtalgos} over three
data sets, the PKDD 2006 Spam Task A~\cite{PKDD06} data set ($k=3$, $d=106780$,
$n=7500$), the School~\cite{School} data set ($k=139$, $d=28$, $n=15362$) and
the Sentiment~\cite{Sentiment} data set ($k=4$, $d=473856$, $n=8000$).  Since
examples in the School data set have real valued labels, a preprocessing was
required to turn those into binary values. Therefore, we assigned a positive
label to those instances whose original score was above the $75$-th percentile
and a negative label to those instances whose original score was below that
threshold. Moreover we rescaled non binary features in the range $[0,1]$. We used a Gaussian kernel for
the School data set and a polynomial kernel for the PKDD 2006 Spam
Task A and Sentiment data sets.

The simple yardstick for our algorithms is the online classifier that runs a
battery of $k$ Perceptrons in parallel with no constraints on the size of their
active sets.  We evaluated the budget multitask algorithms for different sizes
of their budget. In particular, we set $B$ to $25\%$, $10\%$ and $5\%$ of the
size of the active set obtained by the baseline algorithm after a single
training epoch. For {\sc mtbprj} and {\sc mtbprj-2} we set $\eta=0.01$.

Imposing a budget restriction should be detrimental to the overall performance
and it should be even more so when $k$ rather than a single task are to be
processed. On the other hand, a proper multitask formulation should lessen and
ideally negate, this impact. The F-measure values achieved after a single pass
over the training set are reported in Table~\ref{t:exps}. The numbers show that
disregarding multitask information (i.e., $G=D$), and imposing a constraint on
the size of the active set, really negatively affects the performance of budget
multitask algorithms, and this behavior is unsurprisingly shared by all
algorithms.  Even in this case, however, the projection based algorithms tend
to present a relatively better behavior confirming that the projection schemes
are a first step towards a better retention. Specifically, the global
projection step employed by {\sc mtbprj-2} turns out to be particularly
effective, as evidenced by comparing the F-measures obtained on the School data
set, and to a lesser extent on the PKDD 2006 Spam Task A data set, when
$B=5\%$.

It is of course more interesting to see how this decrease in the overall
performance can be offset by taking into account multitask relations. In fact,
setting $G=C$ results in better F-measure values for all three data sets and
for all algorithms. Moreover the differences in the F-measures obtained for the
different choices of G grow larger as the budget size is shrunk to smaller
values. This should not come to a surprise, since when the available memory is
scarce an efficient management policy, which is the main benefit that the
multitask kernel brings to budget algorithms, becomes crucial. It is
particularly surprising that the multitask kernel can be so effective that for
$B=25\%$ all four algorithms match the performance of the baseline and, for
{\sc mtbprj} and {\sc mtbprj-2} this holds true even when $B$ goes down to
$5\%$.

\section{Appendix}
\begin{proofof}{Proposition~3.1}
First, observe that the Laplacian matrix for the augmented graph $G'$ is
\[
L_{G'} = 
\left[ \begin{array}{cc}
    A_G     & -\bone \\
    -\bone^{\top}    & k
\end{array} \right]
\]
where $\bone$ is the vector of all ones. We use~\cite[Theorem~3.3.2]{CM91}
to compute the pseudoinverse of $A_{G'}$
\begin{eqnarray*}
A_{G'}^+ & = & 
\left[ \begin{array}{cc}
    A_G^{-1}-(1+k)^{-1}A_G^{-1}\bone \bone^{\top}A_G^{-2}-(1+k)^{-1}A_G^{-2}\bone \bone^{\top}A_G^{-1}     & -(1+k)^{-1}A_G^{-2}\bone \\
    -(1+k)^{-1}\bone^{\top}A_G^{-2} & 0
\end{array} \right] \\
&& 
+\frac{\bone^{\top} A_G^{-3} \bone}{(1+k)^2} 
\left[ \begin{array}{cc}
    A_G^{-1}\bone \bone^{\top}A_G^{-1} & A_G^{-1}\bone \\
    \bone^{\top}A_G^{-1} & 1
\end{array} \right]
\end{eqnarray*}
Denoting with $\be_i$ the $i$-th standard vector of size $k$ and observing that
$A_G \bone=(I+L_G)\bone=1+0$ since $L_G$ is a Laplacian matrix, we have, for
all $i=1,\dots,k$ and $j=1,\dots,k$
\begin{eqnarray}
\nonumber
  (A_{G'}^+)_{i,j} 
  & = & 
  \be_i^{\top} A_{G'}^+ \be_j \\
\nonumber
  & = & 
  \be_i^{\top}A_G^{-1}\be_j-\frac{2}{1+k}+\frac{k}{(1+k)^2} \\
  & = &
  (A_G^{-1})_{i,j}-\frac{2+k}{(1+k)^2} 
\label{eq:entrydef}
\end{eqnarray}
Finally, by~\cite[Theorem~7]{GX04}
it holds that
\begin{equation}
  \label{eq:th7gx04}
  A_{G'}=-\frac{1}{2}\left[R_{G'} - \frac{1}{1+k}\left(R_{G'} \bone \bone^{\top} 
      + \bone \bone ^{\top}R_{G'}\right) 
    + \frac{1}{(1+k)^2}\bone \bone^{\top}R_{G'}\bone \bone^{\top}\right]
\end{equation}
Substituting~(\ref{eq:th7gx04}) back into~(\ref{eq:entrydef}) yields the
desired result.
\end{proofof}

\bibliographystyle{plain}
\bibliography{arxivmtbudget}

\end{document}